\documentclass[a4paper]{article}

\usepackage{INTERSPEECH2021}
\usepackage{amsmath,graphicx}
\usepackage{multirow}
\usepackage{url}
\usepackage{commath}
\usepackage{array}
\usepackage{array,booktabs,caption,ragged2e}
\usepackage{tabularx}
\usepackage{pbox}
\usepackage{amssymb}
\usepackage{bbm}
\usepackage{sansmath}
\usepackage[export]{adjustbox}
\usepackage{hyperref}

\newcommand{\fetimit}{15.1}
\newcommand{\fetimitrecon}{14.2}
\newcommand{\felib}{6.08}
\newcommand{\felibrecon}{5.89}
\newcommand{\fttimitrecon}{16.4}
\newcommand{\ftlibrirecon}{5.72}
\newcommand{\randftlibrirecon}{6.83}
\newcommand{\randfttimitrecon}{17.2}

\title{Speech SimCLR: Combining Contrastive and Reconstruction Objective for Self-supervised Speech Representation Learning}
\name{Dongwei Jiang$^1$, Wubo Li$^2$, Miao Cao$^2$, Wei Zou$^2$, Xiangang Li$^2$}
\address{$^1$ AI Labs, YuanFuDao, Beijing, China
        \\$^2$AI Labs, Didi Chuxing, Beijing, China}
\email{{jiangdongwei@yuanfudao.com}, {\{liwubo, dylancaomiao, zouwei, lixiangang\}@didiglobal.com}}

\begin{document}

\maketitle
\begin{abstract}
    Self-supervised visual pretraining has shown significant progress recently.  
    Among those methods, SimCLR greatly advanced the state of the art in self-supervised and semi-supervised learning on ImageNet.
    The input feature representations for speech and visual tasks are both continuous, so it is natural to consider applying similar objective on speech representation learning.
    In this paper, we propose Speech SimCLR, a new self-supervised objective for speech representation learning. During training, Speech SimCLR applies augmentation on raw speech and its spectrogram. Its objective is the combination of contrastive loss that maximizes agreement between differently augmented samples in the latent space and reconstruction loss of input representation.
    The proposed method achieved competitive results on speech emotion recognition and speech recognition.
\end{abstract}

\section{Introduction}
    Self-supervised learning is gaining popularity recently.
    On one hand, deep learning relies heavily on large amounts of high quality labeled data, but obtaining labeled data is usually very costly and time-consuming.
    On the other hand, self-supervised learning attempts to extract knowledge from unlabeled data. It can potentially discover representations that capture the underlying structure of such data, which helps with the performance and convergence speed of downstream tasks.

    In the area of speech, researchers proposed many self-supervised pretraining algorithms. 
    Wav2vec \cite{Schneider_2019} extracts representation from data using a contrastive loss that requires a true future audio sample to be distinguished from negative samples. Reference \cite{vq-wav2vec, Baevski2019EffectivenessOS, wav2vec2} extend this approach and learn vector quantized (VQ) representations of audio data using a future time-step prediction task with the help of BERT. 
    Autoregressive Predictive Coding (APC) is a reconstruction loss proposed by \cite{Chung_2019, Chung2019GenerativePF}. It got inspiration from language model in NLP and tries to predict unseen future frames based on past frames. The idea is further extended in \cite{DBLP:conf/acl/ChungG20}, which also added the objective of past frame prediction.
    Some other reconstruction methods \cite{Liu2019MockingjayUS, Wang2020UnsupervisedPO, audio_albert, Song2019SpeechXLNetUA, Jiang2019ImprovingTS, mpc_improvement, tera} were motivated by NLP. Most of them applied BERT-style masks on input feature representations and adopted reconstruction objective for representation learning.
    It is worth pointing out that some of the above methods \cite{Wang2020UnsupervisedPO, tera} used augmentation during the training of unsupervised models. These augmentations helped to make learned representation more robust and improved the performance of downstream tasks.

    Also employed augmentation methods, SimCLR \cite{simclr} is a framework for contrastive learning of visual representations. It learns representations by maximizing the agreement between differently augmented views of the same image via a contrastive loss in the latent space. 
    SimCLR improved considerably over previous self-supervised, semi-supervised, and transfer learning methods that require specifically designed architectures in terms of top-1 accuracy for ImageNet. The best result obtained by it can also match its supervised counterparts.
    Inspired by the recent success of SimCLR, we propose Speech SimCLR, a new self-supervised objective for speech representation learning. Speech SimCLR first conducts augmentation on raw speech and its spectrogram, it then learns speech representation by maximizing agreement between differently augmented sample via a contrastive loss in the latent space.
    Contrastive and reconstruction objectives help speech representation learning in different ways. Recent work \cite{Demystifying} claimed that self-supervised representations can be further improved by combining contrastive and reconstruction objectives. We also tested this theory in our work. 

    We made our code and pretrained model publicly available for reproducibility at \nolinkurl{https://github.com/athena-team/athena/tree/simclr}.

\section{Related Work}
\subsection{Data augmentation for Speech Representation Learning}
    There are several previous work that applied data augmentation for speech representation learning. 
    PASE \cite{Pascual_2019} tackled multiple self-supervised tasks jointly using an ensemble of neural networks that cooperate to discover good speech representations. It used four regression workers and three binary discrimination tasks to learn a higher level of abstraction than surface features. 
    In WavAugment \cite{WavAugment}, the author applied a combination of pitch modification, additive noise and reverberation to speech and substantially increased the performance of CPC by 18-22\% on ABX errors on Libri-light \cite{kahn2019libri}.
    DeCoAR \cite{decoar} and TERA \cite{tera} are two reconstruction-based methods that used augmentation similar to SpecAugument \cite{Park2019SpecAugmentAS}. Apart from temporal and channel augmentation, TERA also added magnitude alteration and got better results on downstream tasks with it.
    Our work differs from these work in the augmentation we applied and the objective function we used.

    \subsection{TERA}
    TERA used a multi-target auxiliary task to pre-train transformer encoder on a large amount of unlabeled speech. The model learns through the reconstruction of acoustic frames from its altered counterpart, where a stochastic policy altering along three dimensions: temporal, channel, and magnitude is used.

    For temporal alteration, $T_{num}$ amount of starting locations $I_T$ without replacement are selected. The amount $T_{num}$ is given as the maximum time alteration percentage $P_T$ normalized by the time alteration width $W_T$. For each starting index location in $I_T$, $W_T$ consecutive frames from it are altered according to the following stochastic alteration policy: 1) 80\% of the time, all the selected frames are masked to zero. 2) 10\% of the time, selected frames are replaced with random segments of frames. 3) For the rest 10\% of the time, the frames in original input are left unchanged.

    For channel alteration, values of a block of consecutive channels are randomly masked to zero for all time steps across the input sequence. The block of masked channels is selected by first sampling the width of block $W_c$ from {0, 1, ..., $W_C$} uniformly. 
    Then, a channel index $I_C$ from \{0, 1, ..., $H_x$ - $W_c - 1$\} is sampled, where $H_x$ is the number of channels in input sequence $x$. The channels from $I_C$ to $I_C + W_c - 1$ are those to be masked.

    Magnitude alteration is done by applying sampled Gaussian noise to augment the magnitude of input sequences with a probability $P_N$ . For $P_N$ of the time, a random magnitude matrix $z$ of dimensions $L_x$ and $H_x$ is sampled, which has the same shape as $x$. Each element in $z$ is sampled from the normal distribution 
    $N$
    with zero mean and 0.2 variance. $z$ is then added on top of the real frames of $x$.

    The final objective of TERA is the L1 reconstruction loss computed between input $x$ and network output from TERA.

\section{Speech SimCLR}
\label{sec:pagestyle}
    \begin{figure}[]
        \includegraphics[width=0.47\textwidth, center]{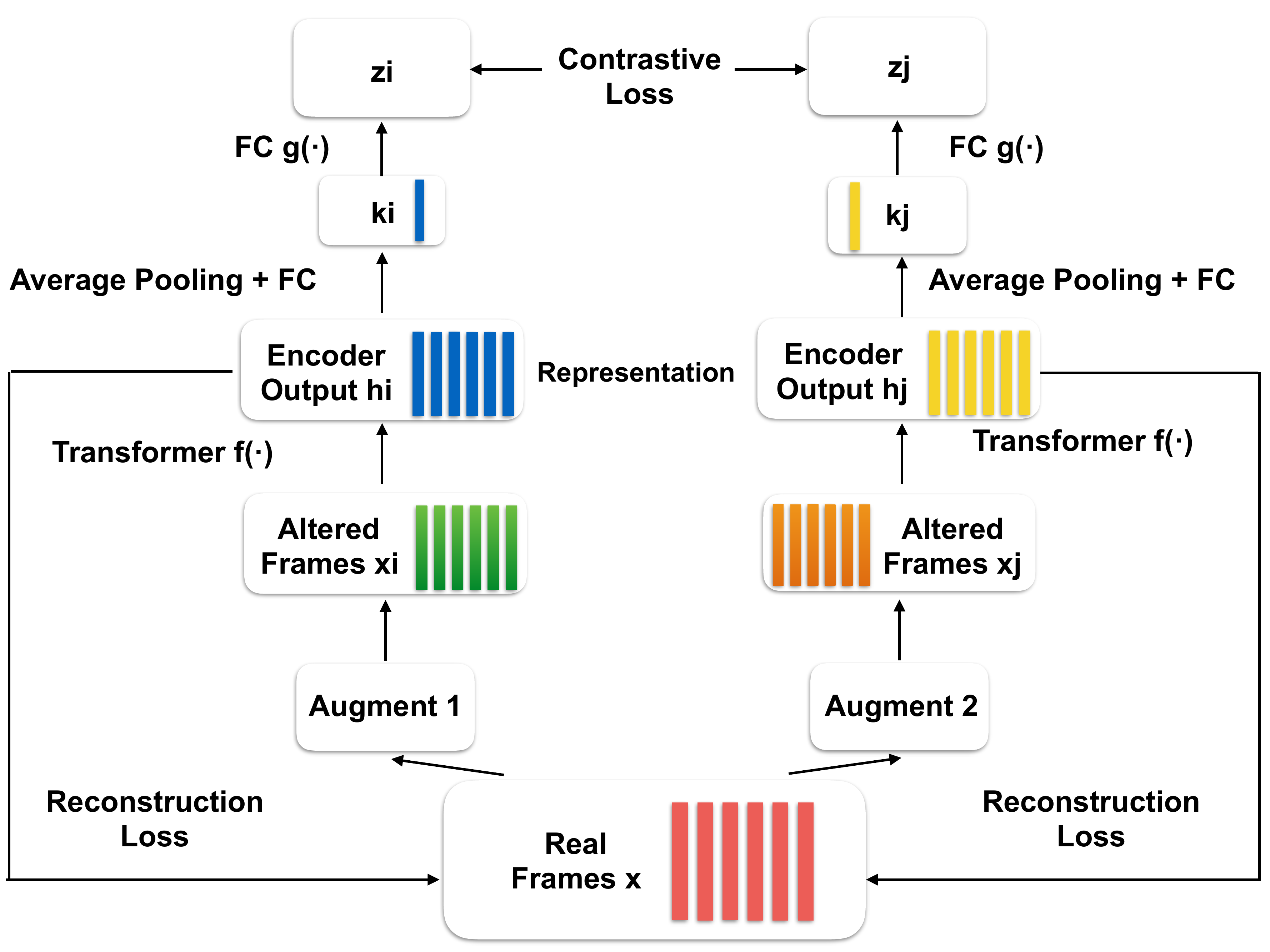}
        \caption{The overall architecture of Speech SimCLR}
        \label{fig:simclr}
    \end{figure}
    
    An illustration of Speech SimCLR is provided in Fig \ref{fig:simclr}. We will discuss each part of Speech SimCLR in detail in the following subsections.

    \subsection{Data Augmentation for Speech}
    For augmentation to a specific speech task, we try to use these methods that would bring variation to input speech but would not affect the final result of downstream tasks.
    For speech recognition and speech emotion recognition, we applied random pitch shift, speed perturbation, room reverberation and additive noise to the original waveform. Time masking and frequency masking are also applied to spectrogram.
    All of our augmentations are conducted using the open source WavAugment tool provided in \cite{WavAugment}.

    \subsection{Model Structure}
    The stochastic data augmentation module transforms any given data example $x$ randomly to two correlated views of the same example $x_i$ and $x_j$. After that, a neural network base encoder $f$ extracts representation vectors from augmented data examples $h_i = f(x_i)$ and $h_j = f(x_j)$. These encoder outputs are the representations that will be used in downstream tasks.
    Transformer based models have been proven to be very effective for various speech tasks \cite{comparative}, so we used transformer layers as the building blocks for the encoder part of Speech SimCLR.

    After encoder, we applied average pooling on encoder output and use a projection head to map it to the space where contrastive loss is applied. Following SimCLR, we also used a non-linear MLP with one hidden layer as the projection head. Specifically, the output of projection head is $z_i$ = $g(avgpool(h_i))$ = $W^{(2)}\sigma(W^{(1)}avgpool(h_i))$ where $\sigma$ is a ReLU non-linearity.

    For contrastive learning to work, distributed training is usually required to get a large batch size \cite{simclr, moco}. During the training of Speech SimCLR, positive pairs are computed in the same device. If BatchNorm is used, the model can exploit local information leakage to improve prediction accuracy because the BatchNorm mean and variance are aggregated locally per device. So we replaced all BatchNorm in our model to LayerNorm as suggested in \cite{Henaff}.
    
    \subsection{Objective Function}

    The contrastive objective function we used in this work is NT-Xent, which achieved the best results compared to other contrastive objectives in SimCLR. 
    For a minibatch of $N$ examples, given a positive pair, the other $2(N - 1)$ augmented examples within a minibatch is treated as negative examples. Let $sim(\textbf{u}, \textbf{v})$ = $\textbf{u}^T\textbf{v}/||\textbf{u}|| ||\textbf{v}||$ denote the cosine similarity between two vectors $u$ and $v$. Then the loss function for a positive pair of examples $(i, j)$ is defined as
    \begin{equation}
        l_{i,j} = -log \frac{exp(sim(z_i, z_j)/\tau)}{\sum_{k=1}^{2N}\mathbbm{1}_{[k \neq i]}exp(sim(z_i, z_k)/\tau)}
    \end{equation}
    where $\mathbbm{1}[k=i]\in$ $\{0, 1\}$  is an indicator function evaluating to 1 iff $k \neq i$ and $\tau$ denotes a temperature parameter. The final loss is computed across all positive pairs, both $(i,j)$ and $(j,i)$, in a mini-batch.
    The temperature of NT-Xent is set to 0.1 throughout our experiments.

    For reconstruction loss, we first adopted time alteration and channel alteration strategy similar to TERA. 
    For temporal alternation, we reduced the time alteration width $W_t$ to 4 frames because in contrastive training, time masking has already been applied.
    Channel alteration width $W_c$ is reduced to 4 channels for the same reason.
    The objective for reconstruction is the L1 loss computed between input and encoder output.

    The final objective for Speech SimCLR then becomes the weighted sum of NT-Xent and reconstruction loss. We gave them equal weights for the experiments of this paper.

\section{Experiments}
\subsection{Data}
    \label{sec:pretraining data}
    
        During training, WavAugument was used to add random pitch shift in the range of -300 to 300 and random speed perturbation in the range of 0.8 to 1.2 to the original audio. Random noise taken from \cite{musan2015} was added to the original audio with SNR between 5-10 db. We also added reverberation with 50\% reverberance, 50\% dumping factor and random room scale in the range of 0-100\% (these are sox-style parameters). Random time mask between 0 to 40 frames and random frequency mask between 0 to 10 mel frequency channels were applied to spectrogram.
        
        For pretraining, all 960 hours of LibriSpeech are used. 80 dimension FBANK is extracted and per-speaker CMVN (cepstral mean and variance normalization) are applied to the features.
        
        IEMOCAP database \cite{iemocap} is used as downstream task for speech emotion recognition. We used the recordings where majority of annotators agreed on the emotion labels and it contains 4 kinds of emotions: angry, happy, sad and neutral. Happy and excited emotions were combined as happy in order to balance the number of samples in each emotion class. The dataset contains 5,531 utterances (1,103 angry, 1,636 happy, 1,708 neutral, 1,084 sad) grouped into 5 sessions. We conducted 5-fold cross validation on IEMOCAP, taking samples from 8 speakers as train and development sets and the ones from the remaining 2 speakers as respective testset.

        For downstream ASR task, the model is fine-tuned on LibriSpeech train-clean-100 set and tested on LibriSpeech test-clean, TIMIT \cite{Lopes2012TIMITAC} and DIRHA \cite{dirha}.
    
    \subsection{Speech Emotion Recognition}
    \label{sec:model_structure}
        The encoder structure of emotion recognition model is similar to TERA, where the encoder is composed of three transformer layers with $num_{layer}=3$, $d_{model}=768$, $d_{ff}=3072$ and $d_{head}=12$. In this task, Speech SimCLR is used as feature extractor. The batch size for Speech SimCLR pretraining is 600 and it was trained for 150 epochs. For fine-tuning, we followed similar model structure as \cite{ruixiong_mpc} and used a transformer encoder with $num_{layer}=12$, $d_{model}=256$, $d_{ff}=2048$ and $d_{head}=4$. The output of transformer encoder is then average pooled and fed into a feed-forward layer.
        Training was done on 4 GPUs with a total batch size of 64 for 25 epochs. We used the Adam optimizer \cite{adam} with warmup schedule \cite{warmup} according to the formula:
            \begin{equation}
                lrate = k * d_{model}^{0.5} * min(n^{-0.5}, n*warmup\_n^{-1.5}),
            \end{equation}
        where $n$ is the step number, $k = 0.5$ and $warmup\_n = 8000$. 
        During evaluating, we averaged from the best 5 checkpoints on the development set during training. The unweighted average of the class-specific recalls (UAR) achieved by the system is used as our metrics.

        As shown in Table \ref{emotion}, Speech SimCLR achieved competitive accuracy on IEMOCAP compared to previous work using other feature.
        OpenAudio contains MuST-C En-De (408 hours) \cite{mustc}, Librispeech (960 hours) and
        ESC-US (347 hours) \cite{esc-us}. Reference \cite{ruixiong_mpc} used self-supervised pretraining with it to boost the performance of IEMOCAP. Despite smaller pretraining dataset size, Speech SimCLR is able to match its performance when used alone and exceed its performance when combined with reconstruction loss.

        \begin{table}[]
            \centering          
            \caption{UAR(\%) comparsion with other methods on IEMOCAP}
            \begin{tabular}{l|l|l|l}
                \toprule
                Methods         &Pretraining Data                                          &Hours   &UAR                \\ \midrule 
                Multi-task Learning \cite{DBLP:conf/icassp/Xia015} & -                     &-  &62.5 \\
                Autoencoder \cite{DBLP:conf/icassp/NeumannV19}  &Tedlium \cite{TED-LIUM}    &207  &59.5 \\
                MPC \cite{ruixiong_mpc}                    &OpenAudio \cite{ruixiong_mpc}  &1715   &64.9 \\
                Speech SimCLR           &LibriSpeech                                  & 960   & 64.3                                   \\ 
                Speech SimCLR + Recon           &LibriSpeech                           &960   & 65.1\\ \bottomrule

            \end{tabular}
        \vspace{-0.5\baselineskip}
        \label{emotion}
        \end{table}

    \subsection{Speech Recognition}
    \label{sec:pretraining for HKUST and AISHELL-1}
        Speech SimCLR is used for both feature extraction and fine-tuning in ASR experiments. For feature extraction, we conducted experiments with the Hybrid DNN/HMM ASR modeling setup implemented inside the PyTorch-Kaldi \cite{pytorch-kaldi} toolkit. We followed the best model structure and setup as TERA, namely TERA base + liGRU + LM Rescore, the baseline WER trained with 100 hours LibriSpeech is 6.2. We did not use any data augmentation for a fair comparsion with TERA \cite{tera}.
        The batch size for Speech SimCLR pretraining is 600 and it was trained for 150 epochs. We compared our results with other work that used a similar setup and listed results in Table \ref{feature-extraction}. 
        On LibriSpeech test-clean, Speech SimCLR achieved lower error rates than other pretraining methods, even when more fine-tuning data is used like in \cite{vq-wav2vec}. PER with Speech SimCLR on TIMIT is only worse than \cite{Song2019SpeechXLNetUA}, which used a pool of Librispeech, TED-LIUM release2 \cite{TED-LIUM} and WSJ-si284 corpora \cite{wsj} as pretraining data.
        It is also worth mentioning that FBANK is considered to be a worse feature than FMLLR in TERA. With the help of reconstruction objective, Speech SimCLR is able to achieve better results than TERA despite that.
        
        \begin{table}[]
            \centering          
            \caption{Error Rate comparsion (\%) with other pretraining methods as feature extraction. WER is reported for LibriSpeech test-clean while PER is reported for TIMIT}
            \begin{tabular}{l|l|l}
                \toprule
                Datasets            &Model                      &Error Rate                 \\ \midrule     
                \multirow{7}{*}{LibriSpeech}    &Bidir-CPC \cite{Kawakami2020LearningRA}    &9.41  \\
                                    &vq-wav2vec \cite{vq-wav2vec}   &6.20    \\
                                    &wav2vec-large \cite{decoar} &6.92   \\
                                    &DeCoAR \cite{decoar}   & 6.10   \\
                                    &TERA + FMLLR \cite{tera}                               &6.05                   \\ 
                                    &Speech SimCLR                 &\felib                   \\ 
                                    &Speech SimCLR + Recon               &\felibrecon                  \\ \midrule
                \multirow{5}{*}{TIMIT}              &wav2vec \cite{Schneider_2019}   &15.6  \\
                                     &TERA + FMLLR \cite{tera}                      &14.5                       \\ 
                                     &Speech XLNET \cite{Song2019SpeechXLNetUA} &13.3 \\
                                    &Speech SimCLR              &\fetimit                       \\ 
                                    &Speech SimCLR + Recon              &\fetimitrecon                      \\ \bottomrule

            \end{tabular}
            \vspace{-0.5\baselineskip}
            \label{feature-extraction}
        \end{table}

        Because our model is written in Tensorflow, it is not possible to conduct fine-tuning experiments using the same setup as feature extraction. Instead, we used end-to-end attention-ctc joint training framework for ASR fine-tuning like the setup in \cite{mpc_improvement}. For encoder, the parameters we used are $num_{layer}=6$, $d_{model}=512$, $d_{ff}=1280$ and $d_{head}=4$, while for decoder, the parameters we used are $num_{layer}=3$, $d_{model}=512$, $d_{ff}=1280$ and $d_{head}=4$. These parameters are carefully chosen so the total number of parameters will be close to TERA base plus ligru. For fine-tuning experiments with Speech SimCLR, a source sequence X is first fed into a prenet that consists of two-layer CNN with 256 channels, stride size 2 and kernel size 3 and transformed to subsampled sequence $X_{0}$ $\in$ $\mathbb{R}^{n^{sub} \times d^{attn}}$ prior to being fed into the encoder.

        During pretraining, because downsampling is applied in encoder, we used a bigger batch size of 1200 and also trained for 150 epochs.
        The setup for fine-tuning and decoding is similar to the one used in \cite{mpc_improvement}. Except that we also applied speed perturbation, additive noise and SpecAugument during fine-tuning stage with the same parameters in pretraining.

        We compared fine-tuned Speech SimCLR with randomly initialized ones and other works on LibriSpeech and TIMIT. For Librispeech, as shown in Table \ref{fine-tuning}, Speech SimCLR achieved lower error rates compared to most other pre-training methods. More importantly, Speech SimCLR still provides benefits to downstream tasks even when similar augmentation is used during fine-tuning. Our result is worse than wav2vec 2.0 \cite{wav2vec2}, but we would like to point out the input feature for wav2vec 2.0 is raw waveform points while the input feature of our work is FBANK. The extra parameters (from CNN feature extractor) and calculation gave wav2vec 2.0 an advantage. Also, the model size of wav2vec 2.0 base contains 95M parameters, which is much larger than the model we used (about 30M parameters). Other factors like quantization also contribute to the superior performance of wav2vec 2.0. We believe the findings in this work and wav2vec 2.0 are complementary and can be applied together to get better speech representation learning.

        The results of Speech SimCLR on TIMIT is worse than other work because the size of TIMIT is too small for end-to-end framework to get competitive results. Nevertheless, fine-tuning works better than the baseline (randomly initialized Speech SimCLR without any untranscribed data), showing the importance of pre-training with more data.

        \begin{table}[]
            \centering          
            \caption{Error Rate(\%) comparsion with other pretraining methods when used as fine-tuning. WER is reported for LibriSpeech test-clean while PER is reported for TIMIT}
            \begin{tabular}{l|l|l}
                \toprule
                Datasets                        &Model                          &Error Rate                 \\ \midrule 
                \multirow{4}{*}{LibriSpeech}    &vq-wav2vec \cite{vq-wav2vec}   &4.50  \\
                                                &TERA + FMLLR \cite{tera}       &5.84                   \\ 
                                                &Rand Init Speech SimCLR        &\randftlibrirecon \\
                                                &Speech SimCLR + Recon          &\ftlibrirecon    \\            
                                                &wav2vec 2.0 \cite{wav2vec2}    &3.40 \\ \midrule

                \multirow{3}{*}{TIMIT}          &TERA + FMLLR \cite{tera}       &15.2           \\ 
                                                &Rand Init Speech SimCLR        &\randfttimitrecon \\
                                                &Speech SimCLR + Recon          &\fttimitrecon                      \\ \bottomrule

            \end{tabular}
        \vspace{-0.5\baselineskip}
        \label{fine-tuning}
        \end{table}

    \subsection{Ablation Study}
    \label{sec:streaming models}
        In this section, we conduct an ablation study to analyze the impact of augmentation, batch size, and training epochs for Speech SimCLR. In all experiments below, Speech SimCLR is used as feature extractor.

        To study the impact of augmentations on different target problems, we retrained a number of Speech SimCLR models with LibriSpeech train-clean-100 set discarding one augmentation at a time.
        For fine-tuning datasets, apart from TIMIT and IEMOCAP, we also added DIRHA \cite{dirha}, a challenging speech recognition dataset containing reverberation and dynamic background noise to see how augmentation in Speech SimCLR will affect downstream tasks in different recording conditions. The training for both TIMIT and DIRHA is done using the MLP configuration in PyTorch-Kaldi \cite{pytorch-kaldi} to make training faster.
        The accuracies of Table \ref{augment} show that no augmentation is dispensable and the best results are achieved when all augmentations are present.
        For individual augmentations, additive noise and speed perturbation are generally the most crucial ones for speech recognition.
        Room reverberation is not so useful for TIMIT and IEMOCAP, but it is really important for DIRHA. We believe its because DIRHA dataset is also artificially contaminated with room reverberation.
        As for emotion recognition, pitch shift and frequency mask are the most important augmentations. We believe its because these two augmentations have the biggest influence on pitch estimation, which is closely related to emotion. 

        \begin{table}[]
            \centering          
            \caption{Baseline accuracy(\%) and absolute accuracy reduction(\%) of TIMIT, DIRHA and IEMOCAP using different augmentation setup for Speech SimCLR. The minus sign in the leftmost column means discarding this augmentation}
            \begin{tabular}{l|l|l|l}
                \toprule
                \multirow{2}{*}{Augmentation}           &\multicolumn{2}{c|}{ASR}           &\multicolumn{1}{c}{Emotion}                            \\ 
                &TIMIT          &DIRHA   &IEMOCAP                           \\ \midrule
                All  &82.7  & 72.5  & 63.7  \\
                - Noise         & -0.7  &-1.3    & -0.5                         \\ 
                - Speed Perturbation & -0.5 &-0.7    & -1.2                         \\ 
                - Time Mask         & -0.2  &-0.3    & -0.3                         \\ 
                - Freq Mask         & -0.2  &-0.4    & -2.4                         \\ 
                - Room Reverb   & -0.0      &-1.2    & -0.2                         \\ 
                - Pitch Shift & -0.2        &-0.2    & -1.9                         \\ \bottomrule

            \end{tabular}
        \vspace{-0.5\baselineskip}
        \label{augment}
        \end{table}

\begin{figure}[]
            \includegraphics[width=0.35\textwidth, center]{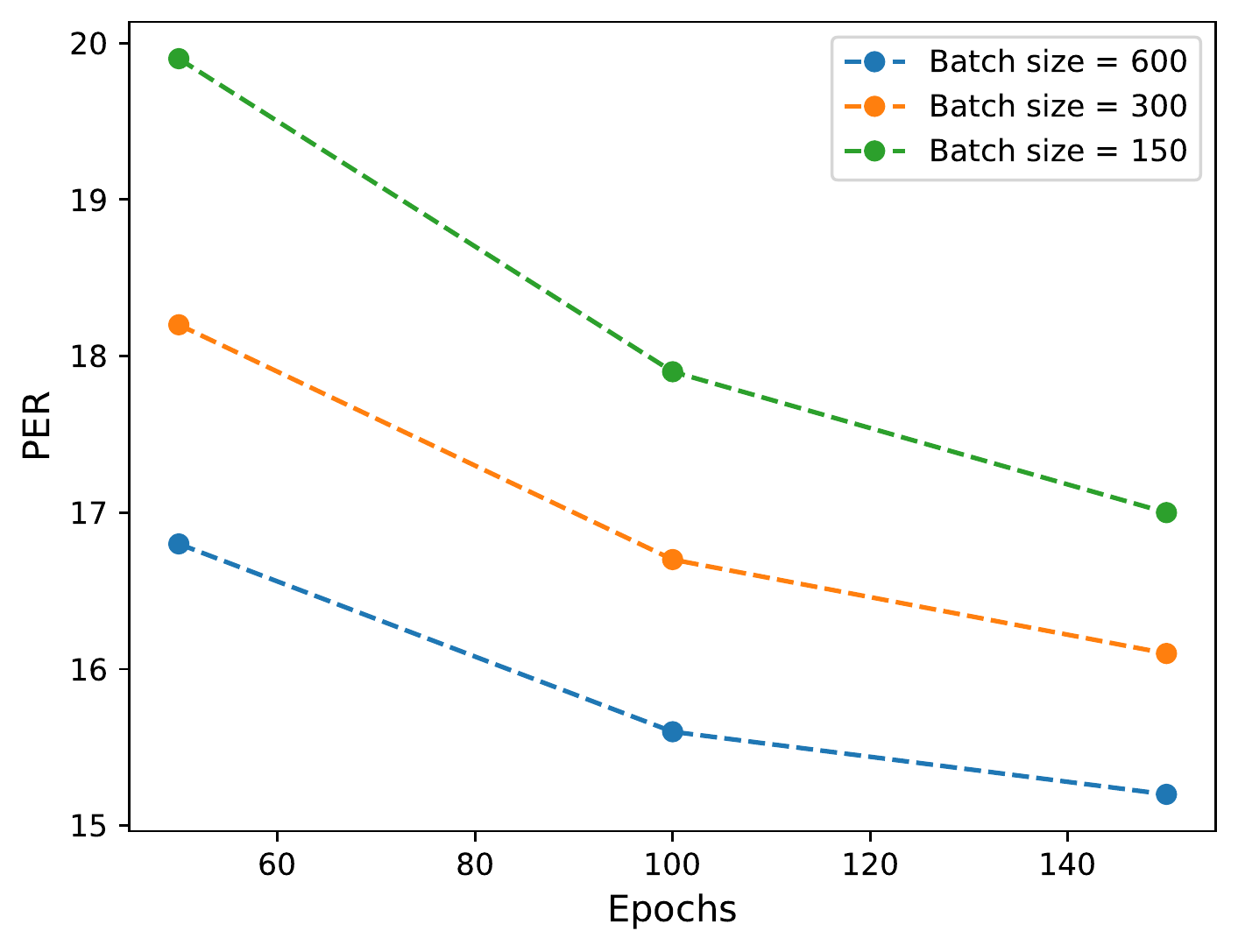}
    
            \caption{PER on TIMIT using LibriSpeech train-clean-100 as pre-training data with different epochs and batch size}
            \label{fig:batch_epoch}
        \end{figure}

        We also evaluated the effect of batch size and training epochs for Speech SimCLR. These experiments are conducted with the whole 960 hours LibriSpeech as pretraining data and LibriSpeech train-clean-100 as fine-tuning data. As shown in Fig \ref{fig:batch_epoch}, using larger batch size and training for more epochs does bring benefits to downstream tasks. We find that, when the number of training epochs is small (e.g. 50 epochs), larger batch sizes have a significant advantage over the smaller ones. With more training steps/ epochs, the gaps between different batch sizes decrease or disappear. The finding here echoes with the conclusion from \cite{simclr} and shows that larger batch sizes are beneficial for contrastive learning because it provides more negative examples. Training longer has the same effect.

\section{Conclusions}

In this paper, we proposed Speech SimCLR, a new self-supervised speech representation learning objective.
Our experiments suggest that Speech SimCLR is able to learn robust representations for both speech recognition and speech emotion recognition.
We also conducted ablation experiments on how augmentation methods, batch size and training epochs will affect the performance of Speech SimCLR.
Moreover, we pointed out that Speech SimCLR can get a consistent boost with the addition of reconstruction loss, which highlights the importance of combining these two objectives.
In the future, we would like to explore other speech augmentations for Speech SimCLR and try to apply Speech SimCLR to other speech related tasks.

\bibliographystyle{IEEEtran}

\bibliography{mybib}

\begin{thebibliography}{10}
\providecommand{\url}[1]{#1}
\csname url@samestyle\endcsname
\providecommand{\newblock}{\relax}
\providecommand{\bibinfo}[2]{#2}
\providecommand{\BIBentrySTDinterwordspacing}{\spaceskip=0pt\relax}
\providecommand{\BIBentryALTinterwordstretchfactor}{4}
\providecommand{\BIBentryALTinterwordspacing}{\spaceskip=\fontdimen2\font plus
\BIBentryALTinterwordstretchfactor\fontdimen3\font minus
  \fontdimen4\font\relax}
\providecommand{\BIBforeignlanguage}[2]{{%
\expandafter\ifx\csname l@#1\endcsname\relax
\typeout{** WARNING: IEEEtran.bst: No hyphenation pattern has been}%
\typeout{** loaded for the language `#1'. Using the pattern for}%
\typeout{** the default language instead.}%
\else
\language=\csname l@#1\endcsname
\fi
#2}}
\providecommand{\BIBdecl}{\relax}
\BIBdecl

\bibitem{Schneider_2019}
S.~Schneider, A.~Baevski, R.~Collobert, and M.~Auli, ``wav2vec: Unsupervised
  pre-training for speech recognition,'' in \emph{Interspeech}, 2019.

\bibitem{vq-wav2vec}
A.~Baevski, S.~Schneider, and M.~Auli, ``vq-wav2vec: Self-supervised learning
  of discrete speech representations,'' in \emph{{ICLR}}, 2020.

\bibitem{Baevski2019EffectivenessOS}
A.~Baevski, M.~Auli, and A.~Mohamed, ``Effectiveness of self-supervised
  pre-training for speech recognition,'' \emph{CoRR}, vol. abs/1911.03912,
  2019.

\bibitem{wav2vec2}
A.~Baevski, H.~Zhou, A.~Mohamed, and M.~Auli, ``wav2vec 2.0: {A} framework for
  self-supervised learning of speech representations,'' \emph{CoRR}, vol.
  abs/2006.11477, 2020.

\bibitem{Chung_2019}
Y.~Chung, W.~Hsu, H.~Tang, and J.~R. Glass, ``An unsupervised autoregressive
  model for speech representation learning,'' in \emph{Interspeech}, 2019.

\bibitem{Chung2019GenerativePF}
Y.~Chung and J.~R. Glass, ``Generative pre-training for speech with
  autoregressive predictive coding,'' in \emph{{ICASSP}}, 2020.

\bibitem{DBLP:conf/acl/ChungG20}
Y.~A. Chung and J.~R. Glass, ``Improved speech representations with
  multi-target autoregressive predictive coding,'' in \emph{ACL}, 2020.

\bibitem{Liu2019MockingjayUS}
A.~T. Liu, S.~Yang, P.~Chi, P.~Hsu, and H.~Lee, ``Mockingjay: Unsupervised
  speech representation learning with deep bidirectional transformer
  encoders,'' in \emph{{ICASSP}}, 2020.

\bibitem{Wang2020UnsupervisedPO}
W.~Wang, Q.~Tang, and K.~Livescu, ``Unsupervised pre-training of bidirectional
  speech encoders via masked reconstruction,'' in \emph{{ICASSP}}, 2020.

\bibitem{audio_albert}
P.~Chi, P.~Chung, T.~Wu, C.~Hsieh, S.~Li, and H.~Lee, ``Audio {ALBERT:} {A}
  lite {BERT} for self-supervised learning of audio representation,''
  \emph{CoRR}, vol. abs/2005.08575, 2020.

\bibitem{Song2019SpeechXLNetUA}
X.~Song, G.~Wang, Z.~Wu, Y.~Huang, D.~Su, D.~Yu, and H.~Meng, ``Speech-xlnet:
  Unsupervised acoustic model pretraining for self-attention networks,''
  \emph{CoRR}, vol. abs/1910.10387, 2019.

\bibitem{Jiang2019ImprovingTS}
D.~Jiang, X.~Lei, W.~Li, N.~Luo, Y.~Hu, W.~Zou, and X.~Li, ``Improving
  transformer-based speech recognition using unsupervised pre-training,''
  \emph{CoRR}, vol. abs/1910.09932, 2019.

\bibitem{mpc_improvement}
D.~Jiang, W.~Li, R.~Zhang, M.~Cao, N.~Luo, Y.~Han, W.~Zou, and X.~Li, ``A
  further study of unsupervised pre-training for transformer based speech
  recognition,'' \emph{CoRR}, vol. abs/2005.09862, 2020.

\bibitem{tera}
A.~T. Liu, S.~Li, and H.~Lee, ``{TERA:} self-supervised learning of transformer
  encoder representation for speech,'' \emph{CoRR}, vol. abs/2007.06028, 2020.

\bibitem{simclr}
T.~Chen, S.~Kornblith, M.~Norouzi, and G.~E. Hinton, ``A simple framework for
  contrastive learning of visual representations,'' \emph{CoRR}, vol.
  abs/2002.05709, 2020.

\bibitem{Demystifying}
Y.~H. Tsai, Y.~Wu, R.~Salakhutdinov, and L.~Morency, ``Demystifying
  self-supervised learning: An information-theoretical framework,''
  \emph{CoRR}, vol. abs/2006.05576, 2020.

\bibitem{Pascual_2019}
S.~Pascual, M.~Ravanelli, J.~Serr{\`{a}}, A.~Bonafonte, and Y.~Bengio,
  ``Learning problem-agnostic speech representations from multiple
  self-supervised tasks,'' in \emph{Interspeech}, 2019.

\bibitem{WavAugment}
E.~Kharitonov, M.~Rivi{\`{e}}re, G.~Synnaeve, L.~Wolf, P.~Mazar{\'{e}},
  M.~Douze, and E.~Dupoux, ``Data augmenting contrastive learning of speech
  representations in the time domain,'' \emph{CoRR}, vol. abs/2007.00991, 2020.

\bibitem{kahn2019libri}
J.~Kahn, M.~Rivi{\`{e}}re, W.~Zheng, E.~Kharitonov, Q.~Xu, P.~Mazar{\'{e}},
  J.~Karadayi, V.~Liptchinsky, R.~Collobert, C.~Fuegen, T.~Likhomanenko,
  G.~Synnaeve, A.~Joulin, A.~Mohamed, and E.~Dupoux, ``Libri-light: {A}
  benchmark for {ASR} with limited or no supervision,'' in \emph{ICASSP}, 2020.

\bibitem{decoar}
S.~Ling, Y.~Liu, J.~Salazar, and K.~Kirchhoff, ``Deep contextualized acoustic
  representations for semi-supervised speech recognition,'' in \emph{ICASSP},
  2020.

\bibitem{Park2019SpecAugmentAS}
D.~S. Park, W.~Chan, Y.~Zhang, C.~Chiu, B.~Zoph, E.~D. Cubuk, and Q.~V. Le,
  ``Specaugment: {A} simple data augmentation method for automatic speech
  recognition,'' in \emph{Interspeech}, 2019.

\bibitem{comparative}
S.~Karita, X.~Wang, S.~Watanabe, T.~Yoshimura, W.~Zhang, N.~Chen, T.~Hayashi,
  T.~Hori, H.~Inaguma, Z.~Jiang, M.~Someki, N.~E.~Y. Soplin, and R.~Yamamoto,
  ``A comparative study on transformer vs {RNN} in speech applications,'' in
  \emph{ASRU}, 2019.

\bibitem{moco}
K.~He, H.~Fan, Y.~Wu, S.~Xie, and R.~B. Girshick, ``Momentum contrast for
  unsupervised visual representation learning,'' in \emph{CVPR}, 2020.

\bibitem{Henaff}
O.~J. H{\'{e}}naff, A.~Srinivas, J.~D. Fauw, A.~Razavi, C.~Doersch, S.~M.~A.
  Eslami, and A.~van~den Oord, ``Data-efficient image recognition with
  contrastive predictive coding,'' \emph{CoRR}, vol. abs/1905.09272, 2019.

\bibitem{musan2015}
D.~Snyder, G.~Chen, and D.~Povey, ``{MUSAN:} {A} music, speech, and noise
  corpus,'' \emph{CoRR}, vol. abs/1510.08484, 2015.

\bibitem{iemocap}
C.~Busso, M.~Bulut, C.~Lee, A.~Kazemzadeh, E.~Mower, S.~Kim, J.~N. Chang,
  S.~Lee, and S.~S. Narayanan, ``{IEMOCAP:} interactive emotional dyadic motion
  capture database,'' \emph{Lang. Resour. Evaluation}, vol.~42, 2008.

\bibitem{Lopes2012TIMITAC}
J.~S. Garofolo, L.~F. Lamel, W.~M. Fisher, J.~G. Fiscus, and D.~S. Pallett,
  ``Darpa timit acoustic-phonetic continous speech corpus cd-rom. nist speech
  disc 1-1.1,'' \emph{NASA STI/Recon technical report n}, vol.~93, 1993.

\bibitem{dirha}
M.~Ravanelli, L.~Cristoforetti, R.~Gretter, M.~Pellin, A.~Sosi, and M.~Omologo,
  ``The {DIRHA-ENGLISH} corpus and related tasks for distant-speech recognition
  in domestic environments,'' in \emph{ASRU}, 2015.

\bibitem{ruixiong_mpc}
R.~Zhang, H.~Wu, W.~Li, D.~Jiang, W.~Zou, and X.~Li, ``Transformer based
  unsupervised pre-training for acoustic representation learning,''
  \emph{CoRR}, vol. abs/2007.14602, 2020.

\bibitem{adam}
D.~P. Kingma and J.~Ba, ``Adam: {A} method for stochastic optimization,'' in
  \emph{ICLR}, 2015.

\bibitem{warmup}
A.~Vaswani, N.~Shazeer, N.~Parmar, J.~Uszkoreit, L.~Jones, A.~N. Gomez,
  L.~Kaiser, and I.~Polosukhin, ``Attention is all you need,'' in \emph{NIPS},
  2017, pp. 5998--6008.

\bibitem{mustc}
M.~A.~D. Gangi, R.~Cattoni, L.~Bentivogli, M.~Negri, and M.~Turchi, ``Must-c: a
  multilingual speech translation corpus,'' in \emph{NAACL-HLT}, 2019.

\bibitem{esc-us}
K.~J. Piczak, ``{ESC:} dataset for environmental sound classification,'' in
  \emph{ACMMM}, 2015.

\bibitem{DBLP:conf/icassp/Xia015}
R.~Xia and Y.~Liu, ``Leveraging valence and activation information via
  multi-task learning for categorical emotion recognition,'' in \emph{ICASSP},
  2015.

\bibitem{DBLP:conf/icassp/NeumannV19}
M.~Neumann and N.~T. Vu, ``Improving speech emotion recognition with
  unsupervised representation learning on unlabeled speech,'' in \emph{ICASSP},
  2019.

\bibitem{TED-LIUM}
A.~Rousseau, P.~Del{\'{e}}glise, and Y.~Est{\`{e}}ve, ``Enhancing the
  {TED-LIUM} corpus with selected data for language modeling and more {TED}
  talks,'' in \emph{LREC}, 2014.

\bibitem{pytorch-kaldi}
M.~Ravanelli, T.~Parcollet, and Y.~Bengio, ``The pytorch-kaldi speech
  recognition toolkit,'' in \emph{ICASSP}, 2019.

\bibitem{wsj}
D.~B. Paul and J.~M. Baker, ``The design for the wall street journal-based
  {CSR} corpus,'' in \emph{ICSLP}, 1992.

\bibitem{Kawakami2020LearningRA}
K.~Kawakami, L.~Wang, C.~Dyer, P.~Blunsom, and A.~van~den Oord, ``Learning
  robust and multilingual speech representations,'' \emph{CoRR}, vol.
  abs/2001.11128, 2020.

\end{thebibliography}


\end{document}